\def\year{2021}\relax
\documentclass[letterpaper]{article} 
\usepackage[utf8]{inputenc}
\usepackage{aaai21}  
\usepackage{times}  
\usepackage{helvet} 
\usepackage{courier}  
\usepackage[hyphens]{url}  
\usepackage{graphicx} 
\usepackage{booktabs}
\usepackage{amsfonts}
\usepackage{amsmath}
\usepackage{natbib}  
\usepackage{caption} 
\newcommand\bM{\mathbf{M}}

\newcommand\bo{\mathbf{o}}
\newcommand\bp{\mathbf{p}}
\newcommand\bh{\mathbf{h}}
\newcommand\bs{\mathbf{s}}
\newcommand\bt{\mathbf{t}}

\newcommand\eg{\textit{e.g.}}
\newcommand\ie{\textit{i.e.}}

\title{Neural Sentence Ordering Based on Constraint Graphs}
\author {
        Yutao Zhu$^\spadesuit$,
        Kun Zhou$^\heartsuit$,
        Jian-Yun Nie$^\spadesuit$,
        Shengchao Liu$^{\spadesuit\clubsuit}$, 
        Zhicheng Dou$^\diamondsuit$ \\
}
\affiliations {
    $^\spadesuit$ Université de Montréal, Montréal, Québec, Canada \\
    $^\heartsuit$ School of Information, Renmin University of China, Beijing, China \\
    $^\clubsuit$ Mila - Québec Artificial Intelligence Institute, Montréal, Canada \\
    $^\diamondsuit$ Gaoling School of Artificial Intelligence, Renmin University of China, Beijing, China \\
    yutao.zhu@umontreal.ca, francis\_kun\_zhou@163.com, nie@iro.umontreal.ca, \\ liusheng@mila.quebec, dou@ruc.edu.cn
}

\begin{document}
\maketitle
\begin{abstract}
Sentence ordering aims at arranging a list of sentences in the correct order. Based on the observation that sentence order at different distances may rely on different types of information, we devise a new approach based on multi-granular orders between sentences. These orders form multiple constraint graphs, which are then encoded by Graph Isomorphism Networks and fused into sentence representations. Finally, sentence order is determined using the order-enhanced sentence representations. Our experiments on five benchmark datasets show that our method outperforms all the existing baselines significantly, achieving a new state-of-the-art performance. The results demonstrate the advantage of considering multiple types of order information and using graph neural networks to integrate sentence content and order information for the task. Our code is available at \url{https://github.com/DaoD/ConstraintGraph4NSO}.

\end{abstract}

\section{Introduction}
Text coherence is an essential problem in natural language processing (NLP). Coherent texts with well-organized logical structures are much easier for people to read and understand. As one subtask of coherence modeling, sentence ordering~\cite{DBLP:journals/coling/BarzilayL08} aims at learning to reconstruct a coherent paragraph from an unordered set of sentences. This task underlies many downstream applications such as determining the ordering of: concepts in concept-to-text generation~\cite{DBLP:conf/acl/KonstasL12,DBLP:conf/emnlp/KonstasL13}, answer spans in retrieval-based question answering~\cite{DBLP:conf/iclr/YuDLZ00L18}, information from various document in extractive multi-document summarization~\cite{DBLP:journals/jair/BarzilayEM02,DBLP:conf/aaai/NallapatiZZ17}, and events in story generation~\cite{DBLP:conf/acl/FanLD19,DBLP:conf/acl/ZhuSDNZ20}. An example of this task is shown in Table~\ref{tab:eg}.

\begin{table}[t!]
    \centering
    \small
    \begin{tabular}{p{0.7cm}p{6.7cm}}
    \toprule
        Order & Unordered Sentences \\
    \midrule
        (2) & When they arrived they saw some airplanes in the back of a truck. \\
        (3) & The \textbf{kids} had a hard time deciding what to ride first. \\
        (1) & The family got together to go to the fair. \\
        (5) & \textbf{\textit{Finally}} they played the dart game. \\
        (4) & Then \textbf{they played} some games to win prizes. \\
    \bottomrule
    \end{tabular}
    \caption{An example of unordered sentences in a paragraph and  their correct order is on the left.}
    \label{tab:eg}
\end{table}

Early studies on sentence ordering generally use handcrafted linguistic features to model the document structure~\cite{DBLP:conf/acl/Lapata03,DBLP:conf/naacl/BarzilayL04,DBLP:journals/coling/BarzilayL08}.
However, these manual features are strongly domain-dependent and their definition requires heavy domain expertise. Therefore, it is difficult to apply these methods to different domains.
To avoid the burden of hand-crafted features, numerous neural approaches have been proposed recently, which can be roughly categorized into two groups: 
The first group tends to solve this problem by predicting the pairwise sentence order with a classifier then inferring the global order~\cite{DBLP:journals/corr/ChenQH16,DBLP:conf/emnlp/LiJ17,DBLP:conf/acl/PrabhumoyeSB20}. An obvious missing part in these approaches is the global context information, which is complementary to the local order information.
Another group predicts the sentence order sequentially based on contextual sentence representations. For example, typical methods are based on pointer networks~\cite{DBLP:journals/corr/GongCQH16,DBLP:conf/aaai/LogeswaranLR18,DBLP:conf/emnlp/CuiLCZ18,DBLP:conf/aaai/YinMSGSZL20,DBLP:conf/aaai/KumarBKR20}, in which sentences and paragraphs are represented by encoders and the order is predicted by a decoder sequentially. However, local coherence information is not taken into account.
Indeed, both local order information and global context information are useful in this task. 
For the example in Table~\ref{tab:eg}, it is preferable to put sentence $s_4$ after $s_3$ as the pronoun ``they'' in $s_4$ may refer to the noun ``kids'' in $s_3$. However, the word ``they'' also appears in $s_2$. Without the contextual information in $s_1$, it would be hard to decide the order between $s_2$ and $s_3$. On the other hand, given the information in $s_1$ and $s_2$, both $s_3$ and $s_4$ can be contextually coherent candidates for the next sentence. It is still hard to decide which should come first after $s_2$ without knowing the relative order $s_3\prec s_4$.

Therefore, we argue that both local information and global context are crucial to sentence ordering. This raises two problems: 1) how to  \textbf{capture} both the local information and the global context; 2) how to \textbf{utilize} them effectively for sentence ordering.

For the first problem, we observe that the order preference between sentences can be hinted by different types of information. 
For example in Table~\ref{tab:eg}, when we compare sentences $s_3$ and $s_4$, the words ``kids'' and ``they played'' indicate that $s_4$ may immediately follow $s_3$.
While comparing sentences $s_5$ and $s_3$, no obvious information can tell that one sentence immediately should follow another. However,  the word ``finally''  in $s_5$ implies that it should appear at some position after $s_3$. In these examples, the clues we use to place sentences at successive positions or at some distance are different. In general, per writing rules, immediately successive sentences may present some strong intrinsic order information (whether it is causal, about time series, or others), while the same type of information may not be observed between sentences separated at a larger distance (\eg, three sentences apart). In the latter case, hints on more global order information can help. For example, ``finally'' suggests a position towards the end, the sentence containing ``go to fair'' should be placed somewhere before ``played some games'' because of their semantic contents. These examples illustrate the need to consider useful order information at different distances or granularities. Based on these observations, we propose to model sentence order at different distances or granularities, each based on its own features.

For the second problem, we propose to model sentence ordering using graph representation. Graph representation is appealing in that different types of information can be naturally described with it. 
In our task, it is natural to treat each sentence in a paragraph as a node and represent their relation (the relative order) by an edge between them. With the help of graph neural networks (GNNs), the representation of each sentence can aggregate the information from its neighbors, leading to a representation that integrates both local order information and global context information.    
More importantly, as we want to take into account sentence relations at different granularities, GNN provides an appropriate way to fuse such information by computing sentence representations over multiple graphs. 
We expect that such an enhanced representation can help better predict sentence order.

More specifically, we design two phases in our framework. In the first phase, we learn multiple classifiers to judge the relative order between two sentences, each within a given distance (granularity level). 
In the second phase, multiple graphs (called \textit{constraint graphs}) are built based on the order preferences. 
Then, we represent sentences as vectors by an encoder and employ GNNs to update these representations based on the graphs. Finally, the sentence representations are fused to predict their order. 

Our main contributions are three-fold:

(1) We propose to capture sentence order information at different granularities, allowing to cover a wide spectrum of useful information; 

(2) We propose a graph representation for sentence orders and design a novel GNN-based method to fuse local and global information; 

(3) We conduct extensive experiments on five benchmark datasets and our model achieves better performance than the existing state-of-the-art methods. This clearly shows the superior capability of our model for sentence ordering by leveraging different types of useful information. Our ablation analysis also shows the impacts of different modules in our framework.

\section{Related Work}
Traditional methods for sentence ordering often rely on handcrafted linguistic features and domain knowledge. For example, \citet{DBLP:conf/acl/Lapata03} computed transition probabilities between sentences and ordered them by a greedy algorithm. \citet{DBLP:conf/naacl/BarzilayL04} proposed a content model which represents topics as states in a Hidden Markov Model (HMM). Then \citet{DBLP:journals/coling/BarzilayL08} took into account entities and computed entity transition probabilities among sentences. Recently, neural network based methods have shown great capability in sentence ordering. We review two groups of neural approaches most relevant to ours:

\noindent\textbf{Pairwise model}. Pairwise models first predict pairwise sentence order, based on which the entire order is inferred. \citet{DBLP:journals/corr/ChenQH16} investigated various methods to judge the order of a sentence pair, and the final ranking score of a sentence is obtained by summing up all its scores in sentence pairs.
\citet{DBLP:conf/acl/PrabhumoyeSB20} proposed to use a topological sort method to infer the entire order based on the pairwise order. While our framework also considers the relative order between each pair of sentences, it also uses a learning method based on graphs rather than a sort algorithm to infer the entire order.
Furthermore, we consider sentence order within multiple distances, leading to a multi-granular view of sentence order. As we will see in our experiments, these extensions significantly improve the results of sentence ordering.

\noindent\textbf{Sequence generation model}. This family of models aims to compute better representations for sentences, that encode some order information.
Typical methods are based on pointer network~\cite{DBLP:conf/nips/VinyalsFJ15}, which is a sequence-to-sequence model using attention as a pointer to select successively a member of the input sequence as the output. \citet{DBLP:journals/corr/GongCQH16} first applied such a model to sentence ordering task, where the encoder represents all sentences into vectors and the decoder predicts results by iteratively selecting one sentence from the input sequence. Later on, many extensions have been proposed.
For example, \citet{DBLP:conf/emnlp/CuiLCZ18} refined the encoder by the self-attention mechanism, while \citet{DBLP:conf/ijcai/YinSSZZL19} modeled the co-occurrences between entities in the sentences by an entity transition graph. 
More recently, researchers found that using feed-forward neural network as decoder and training the whole model by a listwise loss can further improve the performance~\cite{DBLP:conf/aaai/KumarBKR20}. Besides, adding supplementary loss functions during the training process is also helpful~\cite{DBLP:conf/aaai/YinMSGSZL20}. Sentence representations in these approaches are improved, since they can incorporate more contextual information. The relations between sentences are implicitly modeled by self-attention mechanism in these methods. Compared to these methods, our framework explicitly learns and captures the relation at multiple granularities, thus sentence representation can be further improved. 

To our best knowledge, This is the first investigation that utilizes 
graphs to represent the relations between sentences in sentence ordering problem\footnote{Related work about graph representations and GNNs is provided in Appendix.}. Our study will show that graph is a powerful and adequate formalism to represent order-related information for the task.

\begin{figure*}[t!]
    \centering
    \includegraphics[width=\linewidth]{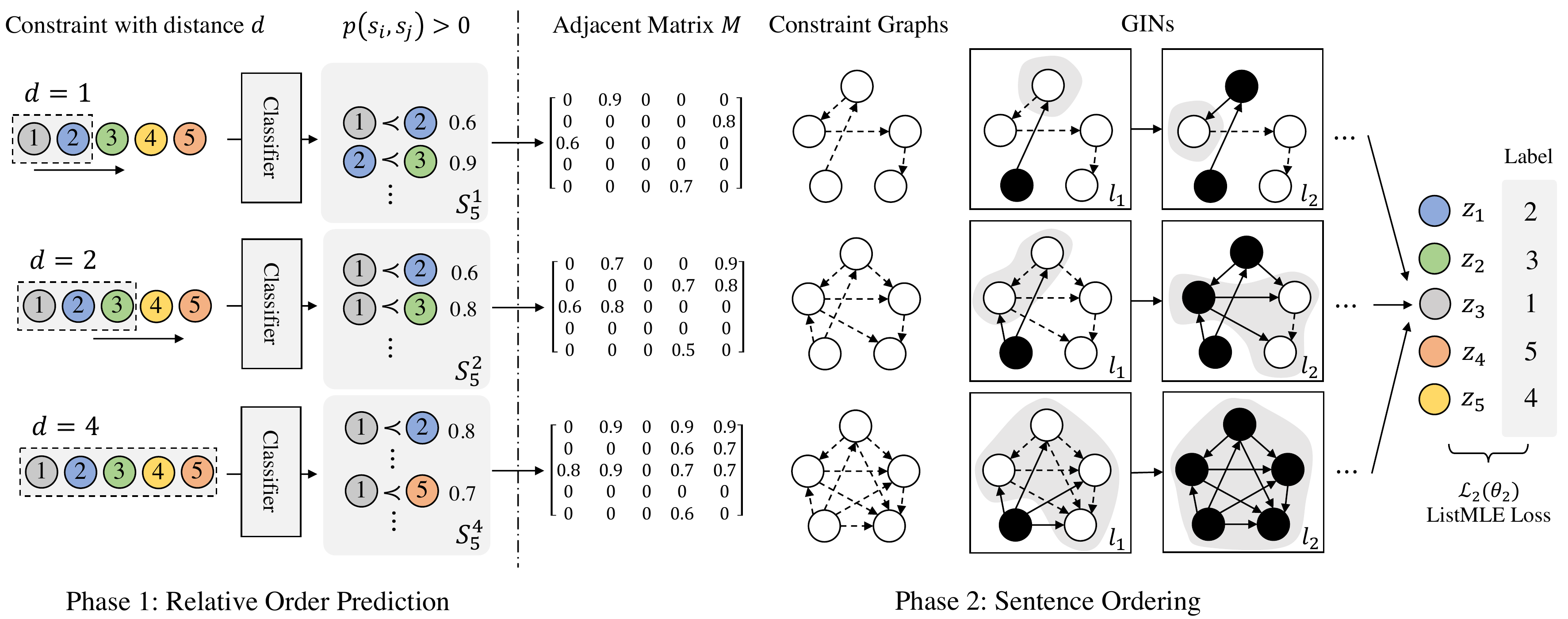}
    \caption{Model architecture. The left part shows the process of the first phase in our method. The example here involves three constraint sets. The second phase is shown on the right side where each constraint set is represented as a constraint graph and integrated into sentence representation by GINs. All sentence representations are fused together to predict the final score of sentences. The second phase of the model is optimized with ListMLE loss.}
    \label{fig:model}
\end{figure*}

\section{Methodology}
\subsection{Task Description}
The sentence ordering task aims at ordering a set of sentences as a coherent text (paragraph). Formally, a set of $n$ sentences with the order $\bo=[o_1, \cdots, o_n]$ can be denoted as $\bp=[s_{o_1},\cdots,s_{o_n}]$. The goal is to find the correct order $\bo^*=[o_1^*, \cdots, o_n^*]$, with which the whole paragraph have the highest coherence probability:
\begin{align}
    P(\bo^*|\bp)>P(\bo|\bp), \forall\, \bo \in \Psi,
\end{align}
where $\bo$ indicates any order of the sentences, and $\Psi$ denotes the set of all possible orders. For example, the order of sentences in Table~\ref{tab:eg} is $[2,3,1,5,4]$, while the correct order $\bo^*$ is $[1,2,3,4,5]$. Following the existing work~\cite{DBLP:journals/corr/ChenQH16,DBLP:conf/aaai/KumarBKR20,DBLP:conf/acl/PrabhumoyeSB20}, this task is framed as a ranking problem, where the model predicts a score for each sentence and the global order is determined by sorting all scores.

\subsection{Model Overview}
As shown in Figure~\ref{fig:model}, our framework contains two phases to capture and to use order information. In the first phase, different from existing methods~\cite{DBLP:journals/corr/ChenQH16,DBLP:conf/acl/PrabhumoyeSB20}, we propose to learn multiple classifiers to predict the relative order at \textit{different distances} between two sentences. Such relative orderings can describe sentence relations in various granularities, thus can provide more comprehensive information for the overall order prediction. This process indeed simulates a common strategy of our human beings to order sentences, \ie, analyzing the relative order between sentence pairs before inferring the entire order. In the second phase, we focus on inferring the overall order based on the previously learned relations. Concretely, we build multiple graphs based on the obtained relations, and employ a GNN on each graph to incorporate the relative ordering information into the sentence representations. Finally, the representations from multiple GNNs are fused together to calculate a final score for each sentence. 

\subsection{Phase 1: Relative Order Prediction}
Given $n$ sentences $[s_{o_1},\cdots,s_{o_n}]$ in any order $[o_1,\cdots,o_n]$, we can create from it a set of constraints ($\mathcal{S}^d_n$) at distance $d$. The set $\mathcal{S}^{d}_n$ describes the relative order between a pair of sentences within distance $d$, which can be described as:
\begin{align}
    \mathcal{S}^d_n=\{s_i\prec s_j \mid 0<j-i\leq d, 1\leq i,j \leq n\}.
\end{align}
For example, assuming the actual order of four sentences is $s_1\prec s_2 \prec s_3 \prec s_4$, if $d=2$, we have five constraints $\{s_1\prec s_2, s_1\prec s_3, s_2\prec s_3, s_2\prec s_4, s_3\prec s_4\}$. 

To obtain $\mathcal{S}^{d}_n$, we build a classifier to predict if the relative order between any two sentences $s_i$ and $s_j$ belongs to this set.
To this end, we fine-tune the Bidirectional Encoder Representations from Transformers (BERT) pre-trained language model~\cite{DBLP:conf/naacl/DevlinCLT19} on each dataset with a multi-layer perceptron (MLP). 
The input is the sequence of tokens of sentence $s_i$, followed by a separator token ``[SEP]'', followed by the sequence of tokens of sentence $s_j$. Then the pooled representation of all the time steps is fed into the MLP and output a probability of $s_i \prec s_j \in \mathcal{S}^d_n$, which is denoted as $p(s_i,s_j)$.
Formally, given $s_i=[w^i_1,w^i_2,\cdots,w^i_u]$ and $s_j=[w^j_1,w^j_2,\cdots,w^j _v]$ containing $u$ and $v$ words respectively, $p(s_i,s_j)$ is computed as:
\begin{align}
    \bt_{ij} &= \text{BERT}([w^i_1,\cdots,w^i_u,[\text{SEP}],w^j_1,\cdots,w^j_v]), \\
    q_{ij} &= \text{Sigmoid}(\text{MLP}(\bt_{ij})), \\
    p(s_i,s_j) &= 
    \begin{cases}
        q_{ij}, & \text{if} \; q_{ij} > 0.5, \\
        0, & \text{else},
    \end{cases} 
\end{align}
where $q_{ij}\in \mathbb{R}$ represents the probability of $s_i \prec s_j \in \mathcal{S}^d_n$. The classifier is trained by the binary cross entropy loss with the ground-truth label $y$:
\begin{align}
    \mathcal{L}_1(\theta_1) = -y \log(q_{ij}) - (1-y)\log(1-q_{ij}).
\end{align}

In our model, we use several distances. Therefore, the above process is repeated several times, each for a given distance, leading to multiple sets of constraints. Notice also that the predicted result for $s_i$ and $s_j$ is not symmetrical because we cannot infer $s_j \prec s_i \in \mathcal{S}^d_n$ from $s_i \prec s_j \notin \mathcal{S}^d_n$. Therefore, $p(s_i, s_j)$ and $p(s_j, s_i)$ should be predicted separately.

\subsection{Phase 2: Sentence Ordering}
The relative orderings are exploited in the second phase to infer the order of all sentences. As suggested by the existing work~\cite{DBLP:journals/corr/ChenQH16,DBLP:conf/aaai/KumarBKR20,DBLP:conf/acl/PrabhumoyeSB20}, the ordering process can be treated as a ranking problem, where the order can be obtained by sorting the scores of sentences. Thus the question is transformed to building better representations for sentences to incorporate the order information.

As shown in Figure~\ref{fig:model}, it is straightforward to represent the determined pairwise order as a graph, in which a sentence corresponds to a node, while the relation between sentences forms an edge.
The advantage of using graphs is that the pairwise relations become explicit, thus provide together a global view of the order information. With the help of GNNs, the connection information (\ie, sentence relation) contained in the edge can be incorporated into the sentence representations. 

More specifically, we build a graph from $\mathcal{S}^d_n$ as follows. First, each sentence $s_i$ is represented by BERT and treated as a node in the graph:
\begin{align}
    \bs_i = \text{BERT}(s_i), i=1,2,\cdots,n,
\end{align}
where $\bs_i\in \mathbb{R}^{768}$ is the representation of the $i^\text{th}$ sentence. Different from the first phase, the input here is a single sentence. Then, an edge $s_i \to s_j$ with weight $p(s_i,s_j)$ is added to the graph if $p(s_i,s_j)>0$. This process forms an adjacent matrix $\bM$ for the graph:
\begin{align}
    \bM_{i,j} = p(s_i,s_j).
\end{align}
We call such a graph \textit{Constraint Graph}. 

We then apply Graph Isomorphism Networks (GINs) to update the representation of the node by recursively aggregating and transforming representation vectors of its neighboring nodes. 
We choose GIN in this work because it achieves state-of-the-art performance on many graph representation tasks~\cite{DBLP:conf/iclr/XuHLJ19}. Alternatively, it could be replaced by any other GNN model. The node representation is initialized by the BERT representation of sentence (\ie, $\bh_i^0=\bs_i$) and updated in a multi-layer GIN through computation with the adjacent matrix $\bM$. The representation of the $i^\text{th}$ node in the $l^\text{th}$ layer is computed as follows:
\begin{align}
    \bh_i^{l} &= \text{MLP}^{(l)}\big( (1+\epsilon^{(l)})\cdot \bh_i^{l-1} + \sum_{j\in \mathcal{N}_i} \bh_j^{l-1}\big),
\end{align}
where $\bh_i^l \in \mathbb{R}^{m}$, $l\in\{1,2,\cdots,L\}$, $\epsilon$ is a hyperparameter which controls the weights of centered node when doing aggregation  in GNN, and $\mathcal{N}_i$ is the set of neighbor nodes of $i^\text{th}$ node.
Then, the node representations from all layers are concatenated and fused together as:
\begin{align}
    \bh_i = \text{MLP}([\bh_i^0; \bh_i^1; \cdots; \bh_i^L]),
\end{align}
where $[;]$ is concatenation operation.
It is worth noting that the number of layers $L$ should be carefully tuned according to the distance $d$ used in the first phase. We will discuss about this in more detail later.

The above process is applied to the multiple constraint graphs of different distances.
Assuming $k$ constraint graphs, then $k$ GINs with different numbers of layers ($L_1,\cdots,L_k$) compute their own representations for each node. The representations from different GINs will then be fused into an augmented representation $\tilde{\bh}_i$: 
\begin{align}
    \tilde{\bh}_i = \text{MLP}([\bh_i^{(1)}; \cdots; \bh_i^{(k)}]),
\end{align}
where $\tilde{\bh}_i\in \mathbb{R}^{m}$ and $\bh_i^{(k)}$ is the sentence representation in the $k^\text{th}$ GIN.

Finally, we compute a score for each node (\ie, sentence) through an MLP:
\begin{align}
    z_i = \text{MLP}\big(\text{ReLU}(\tilde{\bh}_i)\big),
\end{align}
which determines the order of sentences in the paragraph.

We apply ListMLE~\cite{DBLP:conf/icml/XiaLWZL08} as the objective function, which is a surrogate loss to the perfect order 0-1 based loss function. Given a corpus with $N$ paragraphs, the $i^\text{th}$ paragraph with $n_i$ unordered sentences is denoted by $\bp_i=[s_1,\cdots,s_{n_i}]$. Assume the correct order of $\bp_i$ is $\bo_i^*=[o_1^*,\cdots,o_{n_i}^*]$, then ListMLE is computed as:
\begin{align}
    \mathcal{L}_2(\theta_2) &= -\sum_{i=1}^{N}\log f(\bo_i^*|\bp_i), \\
    f(\bo_i^*|\bp_i) &= \prod_{j=1}^{n_i}\frac{\exp(z_{o_j^*})}{\sum_{k=j}^{n_i}\exp(z_{o_k^*})}.
\end{align}
With ListMLE, the model will assign the highest score to the first sentence and the lowest score to  the last one.

\section{Experiment}
\subsection{Datasets}
Following previous work~\cite{DBLP:conf/emnlp/CuiLCZ18,DBLP:conf/aaai/KumarBKR20}, we conduct experiments on five public datasets. The detailed statistics of these datasets are shown in Table~\ref{tab:sta}.
\begin{itemize}
    \item \textbf{NeurIPS/AAN/NSF abstracts}~\cite{DBLP:conf/aaai/LogeswaranLR18}. These datasets consist of abstracts from NeurIPS, ACL and NSF research award papers. The data are split into training, validation, and test set according to the publication year.
    \item \textbf{SIND captions}~\cite{DBLP:conf/naacl/HuangFMMADGHKBZ16}. This is a visual story dataset. Each story  contains five sentences. 
    \item \textbf{ROCStory}~\cite{DBLP:journals/corr/MostafazadehCHP16}. It is a commonsense story dataset. Each story comprises five sentences. Following~\cite{DBLP:conf/aaai/Wang019a}, we make an 8:1:1 random split on the dataset to get the training, validation, and test set.
\end{itemize}

\begin{table}[t]
    \centering
    \small
    \setlength{\tabcolsep}{1.3mm}{
    \begin{tabular}{lrrrrrr}
    \toprule
        \textbf{Datasets} & \textbf{Min.} & \textbf{Max.} & \textbf{Avg.} & \textbf{Train} & \textbf{Val.} & \textbf{Test} \\
    \midrule
        NeurIPS abstracts & 1 & 15 & 6 & 2,448 & 409 & 402 \\
        AAN abstracts & 1 & 20 & 5 & 8,569 & 962 & 2,626 \\
        NSF abstracts & 2 & 40 & 8.9 & 96,017 & 10,185 & 21,573 \\
        SIND captions & 5 & 5 & 5 & 40,155 & 4,990 & 5,055 \\
        ROCStory & 5 & 5 & 5 & 78,529 & 9,816 & 9,816 \\
    \bottomrule 
    \end{tabular}
    }
    \caption{The statistics of all datasets.}
    \label{tab:sta}
\end{table}
\subsection{Baseline Models} We compare our model with the following methods: 

\noindent\textbf{Traditional methods}: Entity Grid~\cite{DBLP:journals/coling/BarzilayL08}; Window Network~\cite{DBLP:conf/emnlp/LiH14a}. 
These methods use hand-crafted features or neural networks to capture the coherence of text. 

\noindent\textbf{Pairwise models}: Seq2seq~\cite{DBLP:conf/emnlp/LiJ17}; B-TSort~\cite{DBLP:conf/acl/PrabhumoyeSB20}\footnote{Note that the results of B-TSort are slightly worse than those reported in the original paper, because the provided source code does not shuffle the sentence order on test set when applying topological sort algorithm, which artificially improves the results.}. 
These methods predict the relative order between sentence pairs, then infer the entire order.

\noindent\textbf{Sequence generation models}: CNN/LSTM+PtrNet~\cite{DBLP:journals/corr/GongCQH16};  Variant-LSTM+PtrNet~\cite{DBLP:conf/aaai/LogeswaranLR18}; ATTOrderNet~\cite{DBLP:conf/emnlp/CuiLCZ18}; HierarchicalATTNet~\cite{DBLP:conf/aaai/Wang019a}; SE-Graph~\cite{DBLP:conf/ijcai/YinSSZZL19}; ATTOrderNet+TwoLoss~\cite{DBLP:conf/aaai/YinMSGSZL20}; RankTxNet+ListMLE~\cite{DBLP:conf/aaai/KumarBKR20}. 
These architectures adopt CNN/RNN based approaches to obtain the representation for the input sentences and employ the pointer network as the decoder to predict order. The last four methods are extended models based on ATTOrderNet. 
   
\subsection{Evaluation Metrics}
We use Kendall's $\tau$ and Perfect Match Ratio (PMR) as metrics, both being commonly used in previous work~\cite{DBLP:journals/corr/GongCQH16,DBLP:conf/emnlp/CuiLCZ18,DBLP:conf/aaai/LogeswaranLR18,DBLP:conf/aaai/YinMSGSZL20,DBLP:conf/aaai/KumarBKR20}.

\noindent\textbf{Kendall's Tau} ($\tau$): it is one of the most frequently used metrics for the automatic evaluation of text coherence~\cite{DBLP:journals/coling/Lapata06,DBLP:conf/acl/Lapata03, DBLP:conf/aaai/LogeswaranLR18}. It quantifies the distance between the predicted order and the correct order in terms of the number of the inversions. $\tau=1-2I/{n \choose 2}$, where $I$ is the number of pairs in the predicted order with incorrect relative order, and $n$ is number of sentences in the paragraph. This value ranges from -1 (the worst) to 1 (the best).

\noindent\textbf{PMR}: it calculates the percentage of samples for which the entire order of the sequence is correctly predicted~\cite{DBLP:journals/corr/ChenQH16}. $\text{PMR}=\frac{1}{N}\sum_{i=1}^{N}\mathbb{I}\{\hat{\bo}_i=\bo^{*}_i\}$, where $\hat{\bo}_i$  and $\bo^{*}_i$ are the predicted and correct orders for the $i^\text{th}$ paragraph. $N$ is the number of samples in the dataset.

\begin{table*}[t!]
    \centering
    \small
    \begin{tabular}{lcccccccccc}
    \toprule
        & \multicolumn{2}{c}{NeurIPS} & \multicolumn{2}{c}{AAN} & \multicolumn{2}{c}{NSF} & \multicolumn{2}{c}{SIND} & \multicolumn{2}{c}{ROCStory} \\
        \cmidrule(lr){2-3} \cmidrule(lr){4-5} \cmidrule(lr){6-7} \cmidrule(lr){8-9} \cmidrule(lr){10-11} 
        & ${\tau}$ & PMR & ${\tau}$ & PMR & ${\tau}$ & PMR & ${\tau}$ & PMR & ${\tau}$ & PMR \\
    \midrule
        Entity Grid$^\triangle$ & 0.09 & - & 0.10 & - & - & - & - & - & - & - \\
        Window Network$^\triangle$ & 0.59 & - & 0.65 & - & 0.28 & - & - & - & - & - \\
        Seq2seq$^\triangle$ & 0.27 & - & 0.40 & - & 0.10 & - & 0.19 & 12.50 & 0.34 & 17.93 \\
        \midrule
        CNN + PtrNet$^\heartsuit$ & 0.6976$^\dag$ & 19.36$^\dag$ & 0.6700$^\dag$ & 28.75$^\dag$ & 0.4460$^\dag$ & 5.95$^\dag$ & 0.4197$^\dag$ & 9.50$^\dag$ & 0.6538$^\dag$ & 27.06$^\dag$ \\
        LSTM + PtrNet$^\heartsuit$ & 0.7373$^\dag$ & 20.95$^\dag$ & 0.7394$^\dag$ & 38.30$^\dag$ & 0.5460$^\dag$ & 10.68$^\dag$ & 0.4833$^\dag$ & 12.96$^\dag$ & 0.6787$^\dag$ & 28.24$^\dag$ \\
        Variant-LSTM + PtrNet$^\heartsuit$ & 0.7258$^\dag$ & 22.02$^\dag$ & 0.7521$^\dag$ & 40.67$^\dag$ & 0.5544$^\dag$ & 10.97$^\dag$ & 0.4878$^\dag$ & 13.57$^\dag$ & 0.6852$^\dag$ & 30.28$^\dag$ \\
        ATTOrderNet$^\heartsuit$ & 0.7466$^\dag$ & 21.22$^\dag$ & 0.7493$^\dag$ & 40.71$^\dag$ & 0.5494$^\dag$ & 10.48$^\dag$ & 0.4823$^\dag$ & 12.27$^\dag$ & 0.7011$^\dag$ & 34.32$^\dag$ \\
        HierarchicalATTNet$^\diamondsuit$ & 0.7008$^\dag$ & 19.63$^\dag$ & 0.6956$^\dag$ & 30.29$^\dag$ & 0.5073$^\dag$ & 8.12$^\dag$ & 0.4814$^\dag$ & 11.01$^\dag$ & 0.6873$^\dag$ & 31.73$^\dag$ \\
        SE-Graph$^\diamondsuit$ & 0.7370$^\dag$ & 24.63$^\ddag$  & 0.7616$^\dag$ & 41.63$^\dag$ & 0.5602$^\dag$ & 10.94$^\dag$ & 0.4804$^\dag$ & 12.58$^\dag$ & 0.6852$^\dag$ & 31.36$^\dag$ \\
        ATTOrderNet + TwoLoss$^\diamondsuit$ & 0.7357$^\dag$ & 23.63$^\ddag$ & 0.7531$^\dag$ & 41.59$^\dag$ & 0.4918$^\dag$ & 9.39$^\dag$ & 0.4952$^\dag$ & 14.09$^\dag$ & 0.7302$^\dag$ & 40.24$^\dag$ \\
        RankTxNet + ListMLE$^\heartsuit$ & 0.7316$^\dag$ & 20.40$^\dag$ & 0.7579$^\dag$ & 36.89$^\dag$ & 0.4899$^\dag$ & 6.81$^\dag$ & 0.5560$^\dag$ & 13.93$^\dag$ & 0.7215$^\dag$ & 28.30$^\dag$\\
        B-TSort$^\diamondsuit$ & 0.7884 & 30.59 & 0.8064$^\ddag$ & 48.08 & 0.4813$^\dag$ & 7.88$^\dag$ & 0.5632$^\dag$ & 17.35$^\ddag$ & 0.7941$^\dag$ & 48.06$^\ddag$ \\
    \midrule
        Our & \textbf{0.8029} & \textbf{32.84} & \textbf{0.8236} & \textbf{49.81} & \textbf{0.6082} & \textbf{13.67} & \textbf{0.5856} & \textbf{19.07} & \textbf{0.8122} & \textbf{49.52} \\
    \bottomrule
    \end{tabular}
    \caption{Results on five benchmark datasets. $\triangle$ indicates previously reported scores. Models with $\diamondsuit$ are implemented with the provided source code while those with $\heartsuit$ are implemented by ourselves. The numbers here are our runs of the model. Hence, they are slightly different from the numbers reported in the original paper. $\dag$ and $\ddag$ denote significant improvements with our method in t-test with $p < 0.01$ and $p < 0.05$ respectively.}
    \label{tab:re}
\end{table*}

\subsection{Implementation Details}
All models are implemented with PyTorch~\cite{DBLP:conf/nips/PaszkeGMLBCKLGA19} and trained on a TITAN V GPU. 
In the first phase, we employ BERT uncased model~\cite{DBLP:journals/corr/abs-1910-03771} with an MLP to predict constraints. The batch size and learning rate is set as 50 and 5e-5 respectively for all datasets.
In the second phase, sentences are also represented by an uncased BERT model and a dropout layer with rate 0.1 is applied over the representations. Three GINs\footnote{We use the implementation at \url{https://github.com/chao1224/BioChemGNN_Dense}.} with the number of layers $L=\{2,3,5\}$ are used for the three corresponding graphs ($k=3$) obtained in the first phase (denoted as $g_1$, $g_2$ and $g_3$ respectively). The hidden size of all GIN layers is 512 ($m=512$). A ReLU activation function is added between each layer. $\epsilon$ is tuned in $\{0, 0.1, 0.5\}$ and set as 0 according to experimental results on validation set. Batch normalization is applied to avoid overfitting. The batch size is set as 128 for all datasets while the learning rate is set as 1e-4, 5e-4, 6e-4, 4e-4 and 3e-4 for NeurIPS, AAN, NSF, SIND and ROCStory dataset. The maximum number of words in sentences is set as 50 on all datasets, which means sentences containing more than 50 words are truncated while those having less than 50 words are padded. All paragraphs in the datasets are randomly shuffled. To make a fair comparison, all models are tested on the same test set without any other preprocessing operations. The models in both two phases are optimized with AdamW~\cite{DBLP:conf/iclr/LoshchilovH19}.

\subsection{Relationship between distance $d$ and the number of layers $L$}
In our experiments, we observe that the hyperparameter $L$ in the second phase should be selected according to $d$ in the first phase. The larger the distance $d$ is, the fewer the GIN layers are required to obtain good results. This trend can be explained by the coverage of the entire set of sentences in a paragraph as follows.

As illustrated in Figure~\ref{fig:model}, assuming that there are five sentences to be ordered, 
we can see when $d=1$ (the top one), the classifier in the first phase only predict the relative order between two successive sentences. Therefore, in the corresponding constraint graph, the model need to stack four layers (hops) to connect the first sentence to the last sentence. Similarly, if $d=2$, the model need two layers to build the connection; and if $d=4$, any two sentences in the graph are connected, thus only one layer is enough. 

Therefore, to connect any sentence pair within a set of $n_i$ sentences for paragraph $i$, we have the empirical formula:
\begin{align}
    L_i \geq \lceil(n_i-1) / d_i\rceil, \label{eq:l1}
\end{align}
where $\lceil\cdot\rceil$ is the ceiling function. In practice, we have to fix $d$ in order to train classifiers, and we can only implement models with a fixed $L$. Therefore, in our experiments, we use GNNs with $L=\{2,3,5\}$ for all datasets and compute the corresponding distance $d$ as:
\begin{align}
    d_i = \lceil \big(\max(\{n_i\}_{i=1}^{N}) - 1\big) / (L_i - 1) \rceil. \label{eq:l2} 
\end{align}
As a result, the distance $d$ is set as \{14, 7, 4\}, \{19, 10, 5\}, \{39, 20, 10\}, \{4, 2, 1\}, and \{4, 2, 1\} for NeurIPS, AAN, NSF, SIND, and ROCStory dataset respectively. 

To validate the Equation~(\ref{eq:l1}), we design an experiment that use the ground-truth constraint graphs as input to test how many layers are necessary to obtain perfect results. 
The results on SIND validation sets ($n=5$ for all samples) confirm our empirical analysis. The details of this experiment is presented in Appendix.

\subsection{Sentence Ordering Results}
Table~\ref{tab:re} shows the sentence ordering results on all datasets\footnote{The average accuracy of each classifier in the first phase on all datasets is around 80\%, and the detailed results are reported in Appendix.}. Our method significantly outperforms all baseline models on both evaluation metrics.

The improvement is generalized across different topics and  sizes of data. On research paper abstracts datasets, our model outperforms the previous best baseline method by about 1.5\% $\tau$ score and 1.7\% PMR score on NeurIPS and AAN datasets, while the improvement is more than 2.5\% on larger datasets, \ie, NSF. As for two story datasets, our method outperforms the previous state-of-the-art model by around 1.8\% $\tau$ score and 1.3\% PMR score. Interestingly, our method achieves 49.52\% PMR score on ROCStory, meaning that about half of the stories in the test set can be ordered completely right. The performance clearly demonstrates the effectiveness and wide applicability of our proposed method.

B-TSort is a recently proposed pairwise method, which predicts the entire order merely based on the relative order of two sentences at only one distance (as $d=n-1$ in our framework). Compared to it, our method can achieve better performance because multiple sentence relations are considered. Besides, in the second phase, we use neural networks rather than a sorting algorithm (\eg, a topological sort) to infer the entire order, which can effectively fuse the order information from multiple constraint graphs.

According to the results, being equipped with BERT may bring improvements (\eg, RankTxNet+ListMLE on SIND and ROCStory dataset). However, compared with the last two baselines that also use BERT as encoder, our framework still yields better results. This indicates that the better performance we obtained is not merely due to BERT embeddings but also to the way we create representations on top of it.

\subsection{Ablation Study}
We investigate the impact of different modules in our model by an ablation study. The experiments are performed on AAN abstracts and SIND captions dataset. From the results shown in Table~\ref{tab:ab}, we can observe:

(1) The effect of each graph varies on different datasets. Compared with using the graph $g_1$ (\ie, the graph with largest distance), our model with $g_3$ (smaller distance) performs better on AAN  but worse on SIND. However, using more graphs always lead to better results.
This empirically validates our assumption that multiple graphs can indeed capture various types of information between sentences that are useful for sentence ordering.

(2) In our model, the BERT encoder used in the second phase is not fine-tuned during the training process. We also test the performance with fine-tuning BERT. The result shows that fine-tuning BERT can bring further improvements in terms of $\tau$. Nevertheless, the number of parameters becomes significantly larger than without fine-tuning (117M vs. 7M), and the training time is much longer, \eg, about 2.41 times more (917s vs. 381s per epoch) on SIND dataset. This suggests that fine-tuning BERT is a good strategy only if we have the necessary computation power. 

(3) As many variants or extensions of BERT have been proposed recently, we also test our method with RoBERTa~\cite{DBLP:journals/corr/abs-1907-11692} and ALBERT~\cite{DBLP:conf/iclr/LanCGGSS20}. However, according to our experiments, these two models cannot perform well in the first phase. The potential reason is that they both remove the Next Sentence Prediction objective in the pre-training stage, which is very important for the constraint prediction of our framework. When applying them in the second phase, the final results show some slight improvements on AAN dataset. Through this experiment, we see that our method can work with other advanced pre-trained language models.

(4) We mentioned that GIN could be replaced by any GNN model. In this experiment, we replace it by graph convolutional networks (GCNs)~\cite{DBLP:conf/iclr/KipfW17} which are also widely used in many tasks. The results  with GCNs are slightly worse compared to the model with GINs, but are still competitive. This shows that other GNN models could also be used on our graphs.

\begin{table}[t!]
    \centering
    \small
    \begin{tabular}{lcccc}
    \toprule
         & \multicolumn{2}{c}{AAN} & \multicolumn{2}{c}{SIND} \\
         \cmidrule(lr){2-3} \cmidrule(lr){4-5}
         & $\tau$ & PMR & $\tau$ & PMR \\
         \midrule
         Our ($g_1+g_2+g_3$) & 0.8236 & 49.81 & 0.5856 & 19.07 \\
         \midrule
         Only $g_1$ & 0.8078 & 47.79 & 0.5708 & 17.23 \\
         Only $g_2$ & 0.8170 & 48.82 & 0.5377 & 17.17 \\
         Only $g_3$ & 0.8116 & 48.06 & 0.5276 & 18.32 \\
         $g_1+g_2$ & 0.8217 & 49.31 & 0.5744 & 17.74 \\
         $g_1+g_3$ & 0.8196 & 49.54 & 0.5837 & 18.34 \\
         $g_2+g_3$ & 0.8208 & 49.58 & 0.5539 & \textbf{19.56} \\
         with Fine-tuning BERT & \textbf{0.8245} & 49.54 & \textbf{0.5867} & 18.56 \\
         with RoBERTa & 0.8240 & 49.54 & 0.5761 & 17.80 \\
         with ALBERT & 0.8228 & \textbf{49.85} & 0.5820 & 17.27 \\
         with GCN & 0.8206 & 49.24 & 0.5818 & 17.71 \\
    \bottomrule
    \end{tabular}
    \caption{Ablation results on AAN and SIND dataset.}
    \label{tab:ab}
\end{table}
\begin{table}[!t]
    \centering
    \small
    \begin{tabular}{lcccc}
    \toprule
        & \multicolumn{2}{c}{AAN} &\multicolumn{2}{c}{SIND}  \\
        \cmidrule(lr){2-3} \cmidrule(lr){4-5}
        & First & Last & First & Last \\
        \midrule
        Our & \textbf{91.47} & \textbf{80.35} & \textbf{79.80} & \textbf{60.44} \\  
        \midrule
        CNN+PtrNet & 77.60 & 67.24 & 68.23 &47.63 \\
        LSTM+PtrNet & 84.63 & 72.63 & 75.01 & 53.79 \\
        Variant-LSTM+PtrNet & 85.63 & 73.28 & 75.05 & 53.87 \\
        ATTOrderNet & 84.82 & 73.09  & 74.52 & 52.46  \\
        HierarchicalATTNet & 80.35 & 68.71 & 75.07 & 52.53 \\
        SE-Graph & 87.00 & 73.59 & 74.58 & 53.73 \\
        ATTOrderNet + TwoLoss & 86.01 & 73.93 & 75.33 & 53.99 \\
        RankTxNet + ListMLE & 88.11 & 74.27 & 79.51 & 57.25 \\
        B-Tsort & 89.53 & 79.78 & 78.06 & 58.36 \\
    \bottomrule
    \end{tabular}
    \caption{Results for the first and last sentence prediction on AAN and SIND dataset.}
    \label{tab:fl}
\end{table}

\subsection{Further Analysis}
As discussed by~\cite{DBLP:journals/corr/GongCQH16,DBLP:journals/corr/ChenQH16,DBLP:conf/emnlp/CuiLCZ18,DBLP:conf/aaai/KumarBKR20}, the first and last sentence should be paid more attention to due to their crucial positions in a paragraph. We provide the accuracy for these sentences on AAN and SIND datasets in Table~\ref{tab:fl}. Our method gives clearly better results than all baselines. In particular, on SIND dataset, our method achieves the absolute improvement of 2.08\% on the last sentence prediction compared with the previous best method. 

\begin{figure}
    \centering
    \includegraphics[width=\linewidth]{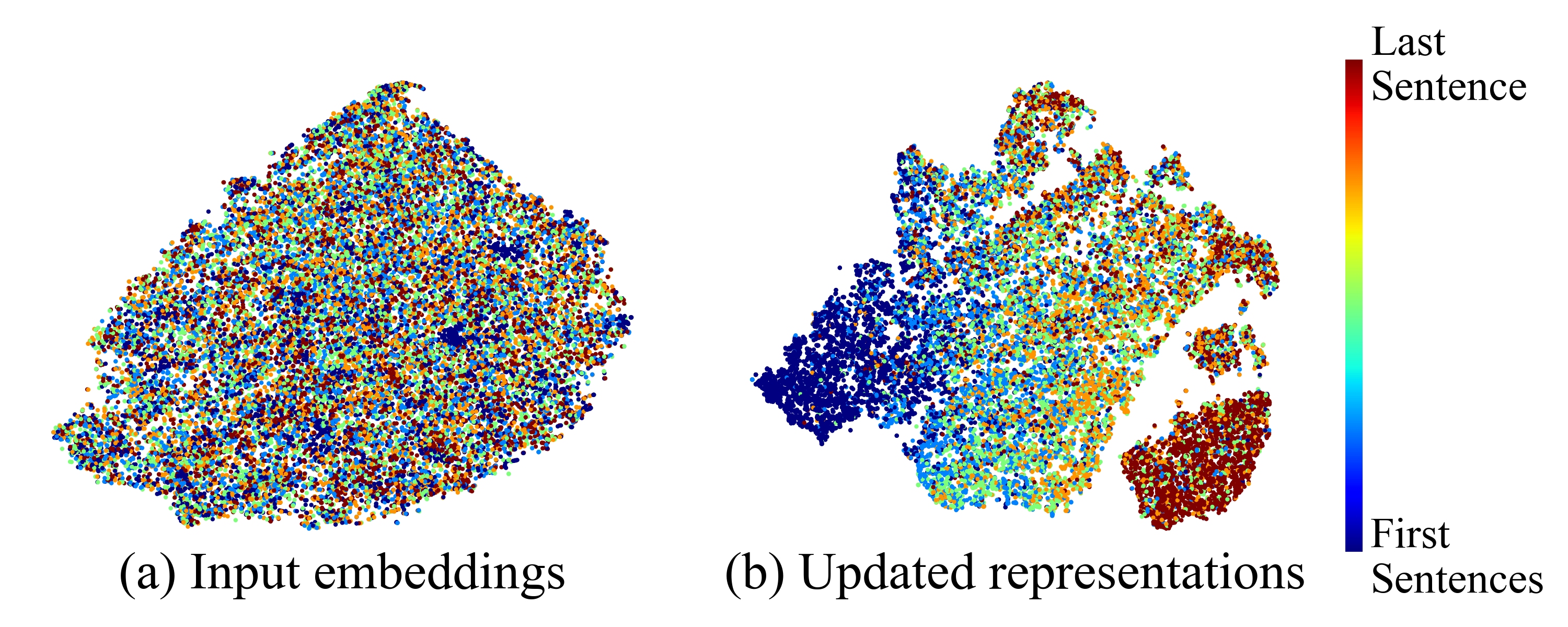}
    \caption{t-SNE embeddings of sentence representations for SIND dataset on sentence ordering task. Colors correspond to the position of the sentences in the original paragraph.}
    \label{fig:sne}
\end{figure}

Following previous work~\cite{DBLP:conf/aaai/LogeswaranLR18,DBLP:conf/aaai/KumarBKR20}, we use t-SNE embeddings to visualize the effect of training on the sentence representations for SIND dataset in Figure~\ref{fig:sne}.
We can clearly see that the updated representations contain more order information than the original BERT embeddings. This is another demonstration that our approach can effectively capture order information into sentence representations. 

\section{Conclusion}
In this paper, we proposed a novel model for sentence ordering which takes into account relative order information at multiple granularities. This is based on the intuition that the order at different distances may rely on different types of information, and all such order information is useful for sentence ordering. We first classified sentence order within different distances and formed multiple constraint graphs from it. 
Then we employed GINs to incorporate the order information into sentence representations, which are finally used to predict the sentence order. 
Our model achieved state-of-the-art performance on five benchmark datasets. This study paves the way for a new research direction on sentence ordering by leveraging different types of information in the form of constraint graphs.

\section{Acknowledgments}
We thank Pan Du for the insightful discussions and the anonymous reviewers for their feedback. This work was supported by National Natural Science Foundation of China No. 61872370 and No. 61832017,  Beijing Outstanding Young Scientist Program NO. BJJWZYJH012019100020098, and Shandong Provincial Natural Science Foundation under Grant ZR2019ZD06.


\clearpage
\section{Appendix}
\subsection{Related work about graph representation and graph neural network}
Graph representation is appealing in the sense that many things can be naturally described with it: as long as the feature has intra-relations, \eg, pixels in images have spatial connection, words in the sentences have temporal correlation, atoms in molecules are connected following the graph topology, and entities in the knowledge graph form a graph in a straightforward way.

The graph neural networks are proposed accordingly for better representation learning, and the core idea is that, for each node, we use message passing operation to exchange information within the $k$-hop neighborhood around it. Successful applications include image classification~\cite{DBLP:journals/corr/abs-2003-00982}, social network analysis~\cite{DBLP:conf/iclr/KipfW17,DBLP:journals/corr/abs-1710-10903}, and molecule property prediction~\cite{DBLP:journals/corr/WuRFGGPLP17,liu2018practical,DBLP:conf/nips/LiuDL19,DBLP:conf/icml/GilmerSRVD17,DBLP:conf/iclr/XuHLJ19}.

\begin{table}[t!]
    \centering
    \small
    \begin{tabular}{lcccccc}
    \toprule
        Dataset & $d$ & $L$ & Acc. of 0 & Acc. of 1 & Overall Acc. \\
    \midrule
        & 14 & 2 & 0.8068 & 0.7599 & 0.7834 \\
        NeurIPS & 7 & 3 & 0.8358 & 0.5143 & 0.7233 \\
        & 4 & 5 & 0.6711 & 0.6791 & 0.6726 \\
    \midrule
        & 19 & 2 & 0.8822 & 0.8761 & 0.8791 \\
        AAN & 10 & 3 & 0.8772 & 0.8850 & 0.8811 \\
        & 5 & 5 & 0.8434 & 0.8625 & 0.8523 \\
    \midrule
        & 39 & 2 & 0.7625 & 0.7133 & 0.7379 \\
        NSF & 20 & 3 & 0.7564 & 0.7634 & 0.7599 \\
        & 10 & 5 & 0.7188 & 0.7243 & 0.7211 \\
    \midrule
        & 4 & 2 & 0.8068 & 0.7599 & 0.7834 \\
        SIND & 2 & 3 & 0.8358 & 0.5143 & 0.7233 \\
        & 1 & 5 & 0.6248 & 0.7439 & 0.6486 \\
    \midrule    
        & 4 & 2 & 0.9008 & 0.9048 & 0.9028 \\
        ROCStory & 2 & 3 & 0.8542 & 0.7693 & 0.8245 \\
        & 1 & 5 & 0.9304 & 0.5647 & 0.8572 \\
    \bottomrule
    \end{tabular}
    \caption{Accuracy of constraints prediction on all datasets. We report the accuracy of label zero (Acc. of 0), the accuracy of label one (Acc. of 1) and the overall accuracy respectively.}
    \label{tab:acc}
\end{table}

\subsection{Accuracy of our method in the first phase}
In the first phase, our method learns a classifier to predict the constraint set with given distance. We report the accuracy of each classifier on the test set in Table~\ref{tab:acc}. As the labels are imbalanced (samples with label zero are more than those with label one), we report the accuracy of each label and the overall accuracy respectively. We can observe that the accuracy is highly related to the dataset and influences the performance in the second phase directly. For example, the accuracy on AAN and ROCStory is higher than that on other datasets. Correspondingly, the performance in the second phase (as shown in Table~\ref{tab:re}) on these two datasets is also better. Moreover, although the accuracy on predicting some of the constraints is not very high, such as $d=4$ on NeurIPS and $d=1$ on SIND, the corresponding graph can still contribute to the final performance on sentence ordering. We believe that any prediction better than random could be useful. Therefore, using multiple graphs is an effective way to improve the sentence ordering performance.

\subsection{Validation for relationship between distance $d$ and the number of layers $L$}
\begin{figure}[t!]
    \centering
    \includegraphics[width=.95\linewidth]{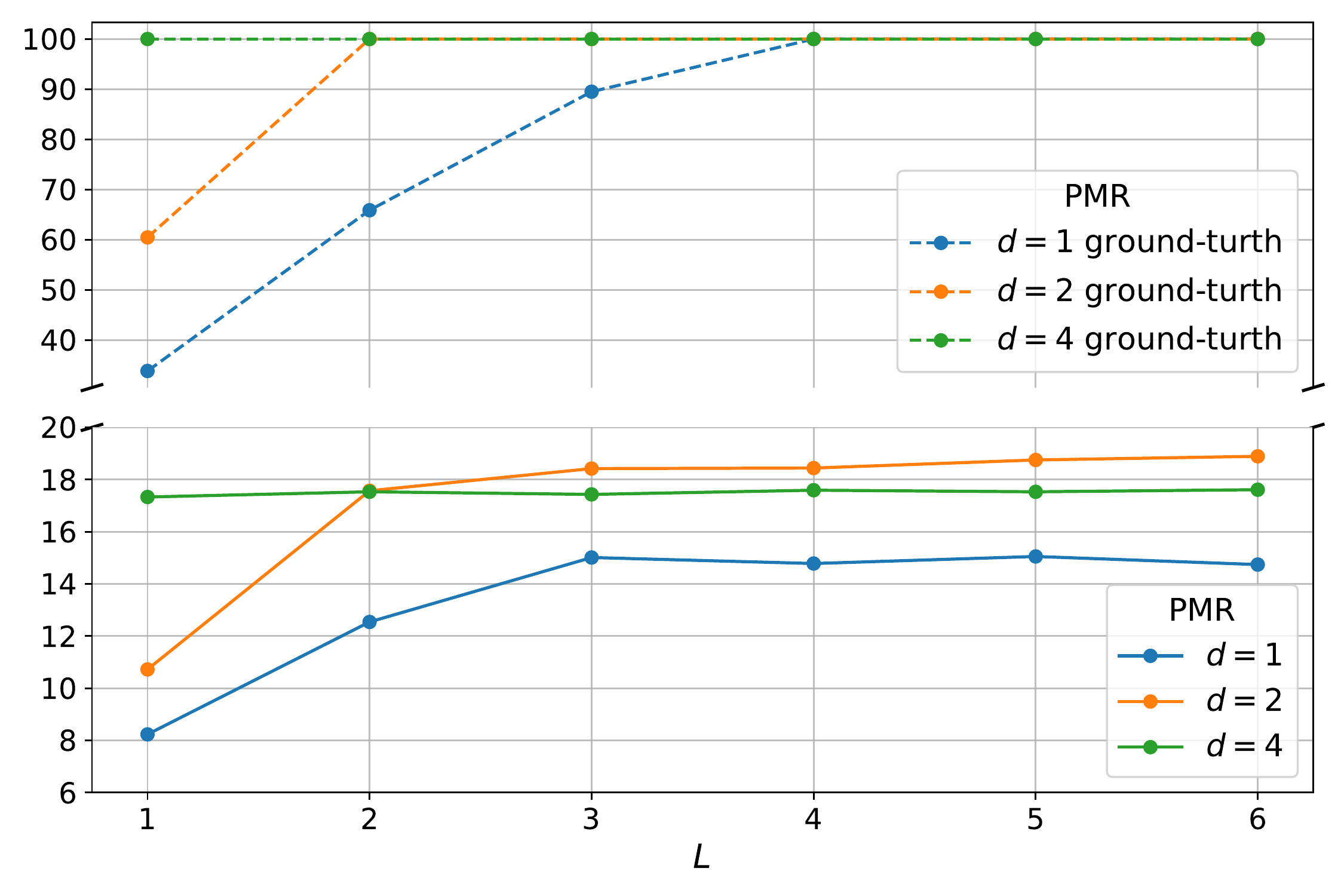}
    \caption{The PMR results on SIND datasets obtained by our method with different distances and different number of layers. The two parts use respectively ground-truth and predicted sentence orders.}
    \label{fig:pmr}
\end{figure}
To validate the Equation~(\ref{eq:l1}), we design an experiment that uses the ground-truth constraint graphs as input to test how many layers are necessary to obtain perfect results. 
The results on SIND validation sets ($n=5$ for all samples) confirm our empirical analysis. 
We observe that the results are consistent with our empirical formula that $d=1$ requires at least four GIN layers ($L\geq4$) to achieve perfect results, while $d=4$ requires only one layer ($L\geq1$). 
Therefore, on this dataset, the necessary setting of $L$ is $\{1,2,4\}$. 
As a reference, the results with predicted constraints are illustrated in the lower part of the figure. According to the results, we find that adding one more layer may slightly improve the performance, thus we emperically set $L$ as $\{2,3,5\}$ on all datasets.

\end{document}